\documentclass[5p,twocolumn]{elsarticle}

\usepackage{lineno}
\modulolinenumbers[5]
\usepackage{stmaryrd}
\usepackage{bbm}
\usepackage{graphicx}
\usepackage{algorithm}
\usepackage{algorithmic}
\usepackage{amssymb}
\usepackage{amsmath}
\usepackage{tabularx}
\usepackage{color}
\usepackage{enumerate}
\usepackage{multirow}
\usepackage{subfigure}
\usepackage{picins}
\usepackage[pagebackref=true,
            breaklinks=true,
            colorlinks,
            urlcolor=blue,  
            citecolor=red,
            bookmarks=false]{hyperref}

\def\ie{i.e.\ }
\def\wrt{wrt.\ }
\def\cf{cf.\ }
\def\etc{etc.\ }
\def\vs{vs.\ }
\def\eg{e.g.\ }
\def\Eg{E.g.\ }
\def\etal{et al.\ }
\def\v{\textbf{v}}
\def\x{\textbf{x}}
\def\pd{\partial}
\def\Loss{\mathcal{L}}
\def\T{\mathcal{T}}
\def\R{\mathbb{R}}
\def\m{\boldsymbol{\mu}}
\def\Sig{\boldsymbol{\Sigma}}
\def\lam{\boldsymbol{\lambda}}

\makeatletter
\def\ps@pprintTitle{%
\let\@oddhead\@empty
\let\@evenhead\@empty
\def\@oddfoot{\reset@font\hfil\thepage\hfil}
\let\@evenfoot\@oddfoot
}
\makeatother

\begin{document}

\begin{frontmatter}

\title{\textbf{Local Higher-Order Statistics (LHS) \\
describing images with statistics of local non-binarized pixel patterns}}

\author[a1,a2]{Gaurav Sharma\fnref{f1}}
\fntext[f1]{GS is currently with Max Planck Institute for Informatics. The majority of the work was
done when GS was with the Universit\'e de Caen Basse-Normandie} 
\ead{grvsharma@gmail.com}
\author[a1]{Fr\'ed\'eric~Jurie}
\ead{frederic.jurie@unicaen.fr}
\address[a1]{GREYC CNRS UMR 6072, Universit\'e de Caen Basse-Normandie, France}
\address[a2]{Max Planck Institute for Informatics, Germany}

\begin{abstract}
 We propose a new image representation for texture categorization and facial analysis, relying on
 the use of higher-order local differential statistics as features. It has been recently shown that
 small local pixel pattern distributions can be highly discriminative while being extremely
 efficient to compute, which is in contrast to the models based on the global structure of images.
 Motivated by such works, we propose to use higher-order statistics of local non-binarized pixel
 patterns for the image description. The proposed model does not require either (i) user specified
 quantization of the space (of pixel patterns) or (ii) any heuristics for discarding low occupancy
 volumes of the space. We propose to use a data driven soft quantization of the space, with
 parametric mixture models, combined with higher-order statistics, based on Fisher scores. We
 demonstrate that this leads to a more expressive representation which, when combined with
 discriminatively learned classifiers and metrics, achieves state-of-the-art performance on
 challenging texture and facial analysis datasets, in low complexity setup. Further, it is
 complementary to higher complexity features and when combined with them improves performance.
\end{abstract}

\end{frontmatter}

\section{Introduction} 

Categorization of textures and analysis of faces under multiple and difficult sources of variations
like illumination, scale, pose, expression and appearance \etc are challenging problems in computer
vision with many important applications. Texture recognition is beneficial for applications
such as mobile robot navigation or biomedical image processing. It is also related to facial
analysis \eg facial expression categorization and face verification (two faces are of same person or
not), as the models developed for textures are generally found to be competitive for face
analysis. Analysis of faces, similarly, has important applications especially in human computer
interaction and in security and surveillance scenarios. This paper proposes a new model for
obtaining a powerful and highly efficient representation for textures and faces, with such
applications in mind.

\begin{figure*}
\centering
\fbox{\includegraphics[width=0.98\textwidth]{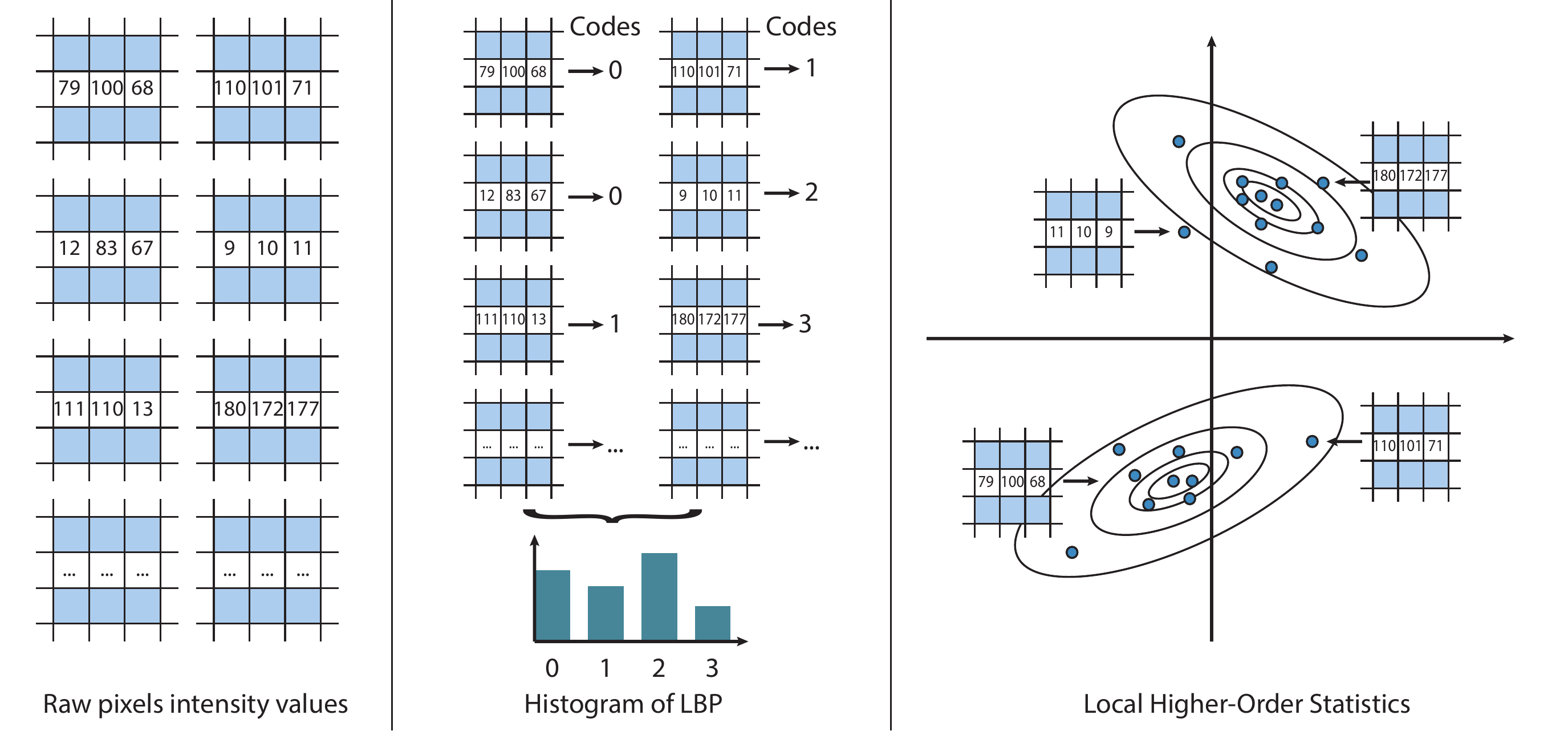}}
\caption{Illustration of the proposed Local Higher-order Statistics (LHS) representation. The
left-hand side of the figure represents a collection of pixel-centered raw pixel intensity values
(image patches). For making the figure simple, we consider only horizontal 2-neighborhood (in
practice we use a 3x3 neighborhood). The middle of the figure shows how these patches can be turned
into LBP codes \cite{Pietikainen:11,Ahonen:PAMI06} -- 4 different codes in this case -- and then
represented as an histogram of LBP. The proposed representation, illustrated on the right-hand side
of the figure, is much richer, as the distribution of the local patches is represented by a Gaussian
Mixture model, encoded as Fisher scores \cite{Jaakkola:NIPS99}.  
\label{fig intro}} 
\end{figure*}

Initial success on texture recognition was achieved by the use of filter banks 
\cite{PuzichaCVPR1997, RubnerICCV1999, Leung:IJCV01,Cula:CVPR01,Zhu:IJCV98}, where the distributions
of the filter response coefficients were used for discrimination. The focus was on evaluating
appropriate filters, selective for edge orientation and spatial-frequencies of variations, and
better capturing the distributions of such filter responses. However, later works \eg by Ojala
\etal \cite{Ojala:PAMI02} and Varma and Zisserman \cite{Varma:CVPR03}, showed that it is possible to
discriminate between textures using pixel values directly (with pixel neighborhoods as small as\
3$\times$3 pixels), discounting the necessity of filter banks. It was demonstrated that despite the
global structure of the textures, very good discrimination could be achieved by exploiting the
distributions of such small pixel neighborhoods. More recently, exploiting such small pixel
neighborhoods or \emph{micro-structures} in textures by representing images with distributions of
local descriptors has gained much attention and has led to state-of-the-art performances for systems
with low complexity, \eg Local Binary Patterns (LBP) \cite{Pietikainen:11,Ahonen:PAMI06},
Local Ternary Patterns (LTP) \cite{Tan:TIP10} and Weber Local Descriptor (WLD) \cite{Chen:PAMI10}.
Most of such local pixel neighborhood based descriptors were shown to be highly effective for
facial analysis \cite{Ahonen:PAMI06, Tan:TIP10} as well. However these methods suffer from important
limitations--the use of fixed hard quantization of the feature space (the space of small pixel
patterns) and the use of heuristics to prune uninteresting regions in the feature space. In addition,
they use histograms to represent the feature distributions. Histograms, or count statistics, are
zeroth order statistics of distributions and thus give a quite restrictive representation.

In contrast, we propose a model that represents images with higher-order statistics of small local
pixel neighborhoods. Fig.~\ref{fig intro} shows an illustration of this representation. We obtain
a data driven soft partition of the feature space using parametric mixture models, to represent the
distribution of the vectors, with the parameters learnt from the training data. Hence, in the
proposed method, the coding of vectors is intrinsically adapted to the data and the computations
involved remain very simple despite the strengths. This helps us avoid the above mentioned
limitations of the previous methods -- (i) instead of a fixed quantization, we learn a data driven,
and hence, adaptive quantization using Gaussian mixture models (GMM), (ii) quantizing using GMM also
avoids any heuristic pruning as any low occupancy region in the feature space will be automatically
ignored by the GMM learning and (iii) learning GMM allows us to use Fisher vectors
\cite{Jaakkola:NIPS99} which are higher-order statistics of the feature distribution. 
We discuss in more detail on this in the following sections. A preliminary version of this work
appeared in Sharma \etal \cite{Sharma:ECCV12}.

We validate the proposed representation by extensive experiments on four challenging datasets: (i)
Brodatz 32 texture dataset~\cite{Valkealahti:PAMI98,Brodatz:66}, (ii) KTH TIPS 2a materials
dataset~\cite{Caputo:ICCV05}, (iii) Japanese Female Facial Expressions (JAFFE)
dataset~\cite{Lyons:AFGR98}, and (iv) Labeled Faces in the Wild (LFW) dataset~\cite{LFWTech}.  
Two dataset, Brodatz-32 and JAFFE, are relatively easier with limited variations while the other
two, KTH TIPS 2a and LFW, are more challenging with realistic high levels of variation in
illumination, pose, expressions etc. We show that using higher-order statistics gives a more
expressive description and lead to state-of-the-art performance in low complexity settings, for the
above datasets. Further, with the challenging LFW dataset as the experimental testbed, we also show
that the proposed representation is complementary to the recent high complexity state-of-the-art
representations.  However, in case of challenging variations, like in LFW, unsupervised approach is
not sufficient and hence we show that when used with supervised metric learning the performance of
the proposed representation improves substantially. When combined with higher complexity methods,
the proposed representation achieves the state-of-the-art performance on the challenging LFW dataset
in the supervised protocol, when no external labeled data is used.

\section{Related works}
\label{sec:relwork}

Texture analysis was initially addressed using filter banks and the statistical distributions of
their responses e.g.\ \cite{PuzichaCVPR1997, RubnerICCV1999,Leung:IJCV01,Cula:CVPR01,Zhu:IJCV98}.
Most of the initial works proposed appropriate directionally and frequency-adapted multiscale filter
banks and/or methods to better capture the statistical distributions of their responses. Later,
Ojala \etal \cite{Ojala:PAMI02} and, more recently, Varma and Zisserman \cite{Varma:CVPR03} showed
that statistics of small pixel neighborhoods, as small as $3\times 3$ pixels, are capable of
achieving high discrimination. This was in contrast to first convolving the local patches with
filter banks and then taking their responses. The success of using raw pixel patches without any
processing discounted the use of filter banks for texture recognition. Since then many methods
working directly with local pixel neighborhoods have been used successfully in texture and face
analysis \eg Local Binary Patterns (LBP) \cite{Pietikainen:11,Ahonen:PAMI06}, Local Ternary Patterns
(LTP) \cite{Tan:TIP10} and Weber Law Descriptor (WLD) \cite{Chen:PAMI10}.

Local pixel pattern operators, such as Local Binary Patterns (LBP) by Ojala \etal
\cite{Ojala:PAMI02}, have been very successful for image description. LBP based image representation
aims to capture the joint distribution of pixel intensities in a local neighborhood as small as
$3\times3$ pixels. LBP makes two approximations, (i) it takes the differences between the center
pixel and its eight neighbors and (ii) then considers just the signs of the differences. The first
approximation lends invariance to gray-scale shifts and the second to intensity scaling. As an
extension to LBP, Local Ternary Patterns (LTP) were introduced by Tan and Triggs \cite{Tan:TIP10} to
add resistance to noise. LTP adds an additional parameter $t$, which defines a tolerance for
similarity between different gray intensities, allowing for robustness to noise. Doing so lends an
important strength: LTPs are capable of encoding pixel similarity information modulo noise using the
simple rule that any two pixels within $\pm t$ intensity of each other are considered similar.  This
is accompanied by a clever split coding scheme to control the size of the descriptor. However, LTP
(and LBP) coding is still limited due to its hard and fixed quantization. In addition, both LBP and
LTP representations usually use the so-called \emph{uniform} patterns: patterns with at most one 0-1
and at most one 1-0 transition, when seen as circular bit strings. It was empirically observed that 
that uniform patterns account for nearly 90\% of all observed pixel patterns in textures, and
hence ignoring the non-uniform patterns leads to large savings in space at negligible loss of
accuracy. Although uniform patterns are beneficial in practice, their use is still a heuristic for
discarding low occupancy volumes in feature space. We will discuss this in more detail in
Sec.~\ref{sec:rel_lbp_ltp}

Owing to the success of the \emph{texton} based texture classification method, \eg Leung and Malik
\cite{Leung:IJCV01}, and the recent success of \emph{bag of words} representation for image
retrieval (by Sivic and Zisserman \cite{Sivic:ICCV03}) and classification (by Csurka \etal
\cite{Csurka2004}) many of the recent methods for texture and face analysis, \eg \cite{Varma:CVPR03,
Liu:ACCV10, Lazebnik:PAMI05, Zhang:IJCV07, Varma:IJCV05, Croiser:IJCV10, Hayman:ECCV04, Xu:IJCV09,
Xu:CVPR10}, use histogram based representations. They first compute a dictionary or codebook of
prototypical vectors, so-called textons or visual words, by clustering large number of randomly
sampled vectors from the training data. The images are then represented as histograms over the
learnt codebook texton assignments. The local vectors are derived in multiple ways, incorporating
different invariances like rotation, view point \etc \Eg \cite{Lazebnik:PAMI05,Zhang:IJCV07}
generate an image specific texton representation from rotation and scale invariant descriptors and
compare them using Earth Movers distance, whereas \cite{Varma:CVPR03, Ojala:PAMI02, Liu:ACCV10,
Varma:IJCV05} use a dictionary learned over the complete dataset to represent each image as
histogram over this dictionary.

In a more recent line of work, Cimpoi \etal \cite{CimpoiCVPR2014} show that traditional image
classification methods when applied to the more challenging textures in the wild scenario give good
performances. They use classic local features such as the Scale Invariant Feature Transform (SIFT)
\cite{Lowe:IJCV04} with different encoding methods, particularly Fisher scores
\cite{Jaakkola:NIPS99}, similar to those employed in the present work. They also evaluate deep
learning \cite{DonahueARXIV2013} based representation and demonstrate their usefulness for the task.
We note that while these method give good performances, they are of much higher complexities than
the proposed method. The proposed method is also complementary to such methods as we will show
empirically later.

We can thus draw a few conclusions from the above mentioned previous works. Modeling distributions
of small pixel neighborhoods (as small as 3$\times$3 pixels) can be quite effective for image
representation \cite{Ojala:PAMI02, Varma:CVPR03, Tan:TIP10}. However, using coarse approximations
(we discuss more on this in Sec.~\ref{sec:rel_lbp_ltp}), as done by most of the previous related
approaches, limits their potential. Finally, the previous methods use low-order statistics,
generally zeroth order counts \ie histograms. This is also limiting as using high-order statistics
can give a more accurate and expressive representation. The main contribution of this paper is
motivated by these observations; we describe small neighborhoods with their higher-order
statistics, without coarse approximations, and show with extensive experimental results that this
leads to a more expressive representation which performs better on challenging benchmark datasets.

In more recent works on facial analysis, deep learning based methods for face
recognition/verification \cite{DuongCVPR2015, HuangCVPR2012, HuCVPR2014, SchroffCVPR2015,
SunCVPR2014, SunCVPR2015, TaigmanCVPR2014, TaigmanCVPR2015, KanCVPR2014, ZhuICCV2013} have gained
much success.  Most of the deep learning based works aim to leverage large amount of data along with
the impressive model capacity of deep networks.  Taigman \etal \cite{TaigmanCVPR2014} showed that
learning a deep convolutional neural network, for predicting thousands of identities using millions
of training images, and then using the output of the penultimate layer of trained network as features
for faces results in very good face verification performance. Alternatively, Huang \etal
\cite{HuangCVPR2012} and Schroff \etal \cite{SchroffCVPR2015} proposed deep architectures to perform
metric learning directly. Kan \etal proposed to handle high variations to pose \cite{KanCVPR2014}
while Schroff \etal \cite{SchroffCVPR2015} propose to use the obtained embeddings to cluster faces
based on identities.

The approach of Martinez \cite{MartinezPAMI2002} is also related to the proposed approach, but the
two are complementary. Martinez \cite{MartinezPAMI2002} proposed to divide face images into small
number of (typically six) local regions and then learn a Gaussian mixture model on PCA compressed
pixel representation of the local face regions. Further, for expression invariant recognition,
Martinez \cite{MartinezPAMI2002} proposed to learn weights on the local regions of the face
corresponding to how important (or, in some sense, invariant) the different regions are, for
recognition of expression variant faces. The proposed method learns the description of an image by
the statistics of very local (3$\times$3 pixel) neighborhoods. Hence the relatively larger in size
(six) local regions in Martinez's approach can be represented by LHS vectors, instead of vectorized
raw pixels compressed with PCA. LHS vectors could be seen as one extreme (highly local)
representation of images, with the PCA based fully global representation at the other extreme.
Martinez's approach can then be seen as striking a balance between the two.

\section{The Local Higher-order Statistics (LHS) model}

We now describe the main contribution of the paper--Local Higher-order Statistics (LHS) model.
LHS intends to represent images by accurately describing the distribution of local pixel
neighborhoods using higher-order statistics. We start with small pixel neighborhoods of 3$\times$3
pixels and model the statistics of their local differential vectors.

\subsection{Local differential vectors.} 
Consider all possible $3\times3$ pixel neighborhoods in an image, \ie 
\begin{equation} 
\v^n = (v_c, v_1,\ldots, v_8)
\end{equation}
where $v_c$ is the intensity of the center pixel and the rest are those of its
8-neighbors. We are interested in exploiting the distribution $p(\v^n | I)$ of the these vectors to
represent the image. Following LBP \cite{Pietikainen:11}, to obtain invariance to monotonic changes
in gray levels, we subtract the value of the center pixel from the rest and obtain the local
differential vectors \ie 
\begin{align}
\v & = (v_1 - v_c, \ldots, v_8 - v_c).
\end{align}
We approximate the distribution of the local pixel patterns with the distribution of the
corresponding differential vectors \ie 
\begin{equation}
p(\v^n | I)  \approx  p(\v | I).
\end{equation}

\subsection{Higher order statistics.}
As the key contribution, we propose to characterize the images using the higher-order statistics of
the differential vectors. We avoid a hard and/or predefined quantization, as used in LBP/LTP, and
use parametric Gaussian mixture model (GMM) to obtain a probabilistic partitioning of the
differential space (\ie the space of all differential vectors). Defining such soft quantization with
mixture model can be equivalently seen as a generative model on the differential vectors. It
allows us to use a characterization method which exploits higher-order statistics \ie
\emph{Fisher score} method proposed by Jaakkola and Haussler \cite{Jaakkola:NIPS99}. 
Fisher scores enables the use of generative modeling with discriminative classifiers. The key idea
is to obtain a fixed length representation of set of vectors, of arbitrary cardinality, by
representing each of the vectors with gradients \wrt the generative model and averaging their
representations (with an iid assumption).
More precisely, given a parametric generative model, a vector $\v$ is characterized by the
gradient of the log likelihood, computed at $\v$, with respect to the parameters of the model. The
Fisher score, for an observed vector $\v$ \wrt a distribution $p(\v | \lam)$, 
is given as,
\begin{equation}
g(\lam, \v) = \nabla_{\lam} \log p(\v | \lam),
\end{equation}
where $\lam$ is the parameter vector.
The Fisher score vector, thus, has the same dimensions as the parameter vector
$\lam$. In the case of a mixture of Gaussian distribution \ie when 
\begin{align}
p(\v | \lam) & = \sum_{k=1}^{N_k} \alpha_k \mathcal{N}(\v|\m_k,\Sig_k),  \\
\mathcal{N}(\v|\m_k,\Sig_k)   & = \frac{1}{\sqrt{(2 \pi)^d |\Sig_k|}} 
            e^{-\frac{1}{2} (\v-\m_k)\Sig_k^{-1} (\v-\m_k)},
\end{align}
the Fisher scores can be computed using the following partial derivatives 
\begin{subequations} \label{eqn:gmm_deriv}
\begin{align}
\frac{\pd \log p(\v|\lam)}{\pd \m_k}        & = \gamma_k \Sig_k^{-1} (\v - \m_k) \\
\frac{\pd \log p(\v|\lam)}{\pd \Sig_k^{-1}} & = \frac{\gamma_k}{2} \left(\Sig_k - (\v - \m_k)^2 \right) \\
\textrm{where, \; \;} \gamma_k              & = \frac{\alpha_k p(\v|\m_k, \Sig_k)}{ \sum_k \alpha_k p(\v|\m_k, \Sig_k)},
\end{align}
\end{subequations}
with the square of a vector being done element-wise.  We have assumed diagonal
$\Sig$, to decrease the number of parameters to be learnt. This amounts to
assuming statistical independence between the variables. Thus, coding vectors using
Eq.~\ref{eqn:gmm_deriv} codes the higher-order, \ie based on the first and second power of $\v$,
statistics of the local differential vectors.
After obtaining the Fisher scores of differential vectors corresponding to every pixel neighborhood
in the image, we compute the image representation as the average of the Fisher scores over all of
them. Here we make an implicit assumption that the vectors were generated iid from the distribution.
This way any image of arbitrary size or equivalently with arbitrary number of vectors is represented
as a vector of length equal to the number of parameters. 

We then perform the following normalizations; first, we normalize each dimension of the image
vector to zero mean and unit variance. To perform the normalization we use training vectors and
compute multiplicative and additive constants to perform whitening per dimension \cite{Bishop:06}.
Second, we perform  power normalization on the image vector $\x$,
\begin{equation}
(x_1,\ldots,x_d) \leftarrow (\textrm{sign}(x_1) \sqrt{|x_1|}, \ldots, \textrm{sign}(x_d) \sqrt{|x_d|}), \label{eqn:n1}
\end{equation}
and finally we do $\ell_2$ normalization of $\x$,
\begin{equation}
(x_1,\ldots,x_d) \leftarrow \left(\frac{x_1}{\sqrt{\sum x^2_i}},\ldots, \frac{x_d}{\sqrt{\sum{x^2_i}}} \right). \label{eqn:n2}
\end{equation}
Perronnin \etal \cite{Perronnin:ECCV10} motivate the power normalization for obtaining a
\emph{de-sparsification} effect, which makes the use of $\ell_2$ distance (and hence the corresponding linear
support vector machine) more appropriate. Similar power normalization has also been shown to be an
\emph{explicit feature map} by Vedaldi and Zisserman \cite{Vedaldi:CVPR10} \ie a mapping which
transforms the vectors to a space where the dot product of the transformed vectors corresponds to
the Bhattacharyya kernel between the original vectors. The whole algorithm, which is remarkably
simple, is summarized in Alg.\ \ref{algo:lhs}. 

Finally, we use the vectors obtained as the representation of the images and employ either
discriminative linear support vector machine (SVM) for supervised classification tasks or
discriminatively learnt Mahalanobis like metrics (detailed below in Sec.~\ref{sec:ml}) for
supervised pair matching, \ie verification, task.

\begin{algorithm}[t]
\begin{algorithmic}[1]
\STATE Randomly sample 3$\times$3 pixels differential vectors $\{\v\in I | I \in \mathcal{I}_{train}\}$
\STATE Learn the GMM parameters $\{\alpha_k, \m_k, \Sig_k| k=1\ldots K\}$ with EM algorithm on $\{\v\}$
\STATE Compute the higher-order statistics, \ie Fisher scores, for $\{\v\}$ using Eq.~\eqref{eqn:gmm_deriv} 
\STATE Compute means $C_{\mu}^i$ and variances $C_{\Sigma}^i$ for each coordinate $i \in \{1,\ldots,d_0\}$
\FORALL{images $\{I\}$}
    \STATE Compute all differential vectors $\v \in I$
    \STATE Compute the Fisher scores for all features $\{\v\}$ using Eq.~\eqref{eqn:gmm_deriv} 
    \STATE Compute the image representation $\x$ as the average score over all features 
    \STATE Normalize each coordinate $i$ as $x^i \leftarrow (x^i - C_{\mu}^i)/C_{\Sigma}^i$
    \STATE Apply normalizations, Eq.~\eqref{eqn:n1} and  \eqref{eqn:n2}
\ENDFOR
\caption{Computing Local Higher-order Statistics}
\label{algo:lhs}
\end{algorithmic}
\end{algorithm}

\subsection{Relation to LBP/LTP.}
\label{sec:rel_lbp_ltp}
We now discuss how LHS can be considered as a generalization of local pattern features.
Consider the Local Binary Patterns (LBP) of Ojala \etal \cite{Ojala:PAMI02}--every pixel is coded as
a binary vector of 8 bits corresponding to its 8 immediate neighbors. Each bit of LBP indicates
whether the corresponding neighboring pixels is of greater intensity than the current pixel or
not. We can thus derive LBP \cite{Ojala:PAMI02} by thresholding each coordinate of our differential
vectors at zero. Hence the LBP space can be seen as a discretization of the differential space into
two bins per coordinate, \ie into the $2^8$ hyperoctants of the 8-dimensional space of local
differential vectors. Similarly, we can discretize the differential space into more number of bins,
with three bins per coordinate i.e.\ $(-\infty, -t), [-t,t], (t,-\infty)$ we arrive at the local
ternary patterns \cite{Tan:TIP10} and so on.  The use of \emph{uniform patterns} (patterns with
exactly one 0-1 and one 1-0 transition), in both LBP/LTP, can be seen as an empirically derived
heuristic for ignoring volumes, in differential space, which have low occupancies, \eg more than
75\% of the hyperoctants for LBP\footnote{Out of the total 256 bins for all possible 8$d$ binary
patterns, 58 bins for uniform patterns and one bin for all the rest of the patterns, are usually
used in LBP}. Thus, the local binary/ternary patterns are obtained with (i) a hard and hand set
quantization of space and (ii) a rejection heuristic derived from empirical observation.  While for LHS such
quantization of space is learnt from data using parametric mixture models, which automatically
adapts itself locally according to the occupancy levels of the space. Hence, in our case the
quantization and rejection is data driven and more general. Moreover, in LBP/LTP the final
representation is based on zeroth order statistics, \ie counts/histograms, while using a data driven
soft quantization allows us to exploit higher-order statistics, as detailed above, for a more
expressive image description.

\section{Discriminative metric learning}
\label{sec:ml}
Recently it has been shown that popular features can be compressed by orders of magnitude by
learning low dimensional projections with a discriminative objective function for the task of pair
matching \ie verification. Such supervised
learning also enhances the discrimination capability of the features upon projection. In the experimental
section, we show the efficacy of the proposed Local Higher-order Statistics (LHS) features when used
with discriminative learning for the challenging task of face verification. In this section, we give
the details of the discriminative metric learning method we use to learn such projection.

\begin{algorithm}[t]
\begin{algorithmic}[1]
\STATE Given: Training set ($\T$), bias ($b$), margin ($m$), learning rate ($r$) 
\STATE Initialize: $L,V \leftarrow$ Whitened PCA of randomly selected training faces $\{\x\}$
\FORALL{$i = 1,\ldots,$\texttt{niters} }
    \STATE Randomly sample a face pair $(\x_i,\x_j, y_{ij})$ from $\T$ 
    \STATE Compute $D_J^2(\x_i,\x_j)$ using Eq.~\ref{eqn:dist} 
    \IF{$y_{ij}(b - D_J^2(\x_i,\x_j)) < m$}
        \STATE $L \leftarrow L - r y_{ij} L (\x_i - \x_j) (\x_i - \x_j)^\top$ 
        \STATE $V \leftarrow V + r y_{ij} V \x_i \x_j^\top$ 
    \ENDIF
\ENDFOR
\caption{SGD for distance learning}
\label{algo:ml}
\end{algorithmic}
\end{algorithm}

Metric learning has recently been a popular topic of research in the machine learning community. While
an exhaustive review of different metric learning methods is out of scope of the paper, we encourage
the interested reader to see an excellent review by Bellet \etal \cite{BelletArxiv2013}.  More closely
related to the present work, metric learning has been successfully applied to the task of face
verification, \ie to predict if two images are of the same person or not. This is different from
face recognition, as the faces may be of person(s) never seen before. The discriminative objectives
used in such methods are based usually on margin maximizing or probabilistic principles
\cite{Guillaumin:ICCV2009, MignonCVPR2012, SimonyanBMVC2013, ChenECCV2012}. Inspired by such works
we now present the method we use to learn a metric using the proposed LHS face representation.

We are interested in learning a `distance' function, for comparing two faces $\x_i$ and $\x_j$,
parameterized by two matrices $L$ and $V$. Our function $D_J(\cdot)$ is a combination of two terms,
first term $D_L(\cdot)$ is the Euclidean distance in the low dimensional space corresponding to the
row space of $L$ and the second $D_V(\cdot)$, is the dot product similarity in another low
dimensional space corresponding to the row space of $V$ \ie
\begin{align}
D_J^2(\x_i, \x_j) & = D_L^2(\x_i, \x_j) - D_V^2(\x_i,\x_j) \label{eqn:dist} \\
D_L^2(\x_i, \x_j) & = \|L \x_i - L \x_j \|^2 \nonumber \\
                  & = (\x_i - \x_j)^\top L^\top L (\x_i - \x_j)\\
D_V^2(\x_i, \x_j) & = \x_i^\top V^\top V \x_j,
\end{align}
where we use the subscript `J' to signify joint Euclidean distance and dot product similarity based
distance. Both the matrices $L$ and $V$ map the original $d_0$ dimensional LHS features to $d \ll
d_0$\footnote{In general the number of rows of $L$ and $V$ can be different. Here, we keep
them the same.} dimensional vectors.  $d$ is a free parameter and is chosen on a per-task basis 
(Sec.\ \ref{sec:lfw_sup}).

We learn the projection matrices $L$ and $V$ by minimizing the following loss function,
\begin{equation}
\label{eqn:loss}
\Loss(\T;L,V) = \sum_{\mathcal{T}} \max \left( 0, m - y_{ij}(b -  D_J^2(\x_i,\x_j) \right)
\end{equation}
where $\T = \{(\x_i,\x_j,y_{ij})\}$ is the provided training set, with pairs of faces $\x_i,\x_j\in
\R^{d_0}$ annotated to be of the same person ($y_{ij}=+1$) or not ($y_{ij}=-1$). Minimization of
this margin-maximizing loss encourages the distance, between pairs of faces of same (different) person, 
to be less (greater) than the bias $b$ by a margin of $m$.

We learn the parameters, \ie $L$ and $V$, with a stochastic gradient descent (SGD) algorithm
with easily calculable analytic gradients outlined in Alg.~\ref{algo:ml}.

\section{Experimental results}
\label{sec:exp}
We now report various experimental results which validate the proposed method. We use four
challenging publicly available datasets of textures and faces and address the challenging tasks of
texture recognition, texture categorization, facial expression categorization and face verification.

In the following, we first discuss implementation details then present the datasets and finally give
the experimental results for each dataset. In the first set of experiments (upto
Sec.~\ref{sec:stateofart}), as our focus is on rich and expressive representation, we use a standard
classification framework based on linear SVM. As linear SVM works directly in the input feature
space, any improvement in the performance is directly related to a better encoding of local regions,
and thus helps us gauge the quality of our representation \vs the competition, with same setup. In
the last part of the experiments (Sec.~\ref{sec:lfw_sup}), we show results with supervised metric
learning, on the LFW dataset, which can also be seen as a discriminatively learned embedding of features.

\subsection{Implementation details.}  
We use only the intensity information of the images and convert color images, if any, to grayscale.
We consider two neighborhood sampling strategies (i) rectangular sampling, where the 8 neighboring
pixels are used, and (ii) circular sampling, where, like in LBP/LTP \cite{Ojala:PAMI02,Tan:TIP10},
we interpolate the diagonal samples to lie on a circle, of radius one, using bilinear interpolation.
We randomly sample at most one million features from training images to learn Gaussian mixture model
of the vectors, using the EM algorithm initialized with k-means clustering. We keep the number of
components as an experimental parameter (Sec.\ \ref{sec:params}). We also use these features to
compute the normalization constants, by first computing their Fisher score vectors and then
computing (per coordinate) mean and variance of those vectors (Alg.\ \ref{algo:lhs}). We use the
average of all the features from the image as the representation for the image. However, for the
facial expression dataset we first compute the average vectors for non overlapping cells of
10$\times$10 pixels and concatenate these for all cells to obtain the final image representation.
Such gridding helps in capturing spatial information in the image and is standard in face analysis
\cite{Feng:PRAI05,Shan:IVC09}. We crop the 250$\times$250 face images to a ROI of $(66,96,186,226)$,
to focus on the face, before feature extraction and do not apply any other pre-processing. Finally,
we use linear SVM as the classifier with the cost parameter $C$ set using five fold cross validation
on the current training set.

In the supervised setting for face verification, we use the metric learning formulation described
above in Sec.~\ref{sec:ml}. We set the bias $b=1.0$, the margin $m=0.2$ and rate $r=0.002$ for all
the experiments. During testing a face pair, we horizontally flip the faces and average the distances
between the four possible pairs of flipped and non-flipped faces. During training, at each SGD
iteration, we randomly select one of the 4 possible flipped/non-flipped pairs for making an update. 

We also combine the proposed LHS with our implementation of Fisher Vectors based on
dense SIFT features (SIFT-FV) \cite{SimonyanBMVC2013, SanchezIJCV2013, Jaakkola:NIPS99}. The
implementation is similar to LHS with the local differential vectors being replaced by dense SIFT
features. We extract SIFT features, using the \texttt{vlfeat} library \cite{vlfeat},  with a step
size of 1 pixel at 5 scales \ie original image and 2 upsampled and 2 downsampled versions
respectively, with a scale difference of $\sqrt{2}$. The SIFT features are compressed to $d_s=64$
dimension using PCA. We use a vocabulary size of $k=16$ and use a spatial grid of $N_c=7\times4$,
giving a feature of dimension $2\times k \times d_s \times N_c 
= 57344$.

\subsection{Baselines.} 
As baselines, we give results with single scale LBP/LTP features generated using the same samplings
as our LHS features, in respective experiments. We use histogram representation over bins of uniform
patterns and add one bin for all the rest of the patterns. We L1 normalize the histograms and
take their square roots and use them with linear SVM. As discussed previously as well, it has been
shown that taking square root of histograms transforms them to a space where the dot product
corresponds to the non linear Bhattacharyya kernel in the original space \cite{Vedaldi:CVPR10}. Thus
using linear SVM with square root of histograms is equivalent to SVM with non linear Bhattacharyya
kernel. Similar square root (\ie power normalization) was also shown to be useful for Fisher scores
\cite{SanchezIJCV2013}.  We note here that our baselines are strong baselines.

\subsection{Texture categorization}
\label{sec:exp_tex}
\noindent \textbf{Brodatz -- 32 Textures
dataset}\footnote{http://www.cse.oulu.fi/CMV/TextureClassification}~\cite{Valkealahti:PAMI98,Brodatz:66}
is a standard dataset for texture recognition. It contains 32 texture classes
\eg\ bark, beach-sand, water,  with 16 images per class. Each of the image is
used to generate 3 more images by (i) rotating, (ii) scaling and (iii) both
rotating and scaling the original image -- note that Brodatz-32
\cite{Valkealahti:PAMI98} dataset is more challenging than original Brodatz dataset
and includes both rotation and scale changes. The images are 64$\times$64 pixels
histogram normalized grayscale images. We use the standard protocol \cite{Chen:PAMI10}, of randomly
splitting the dataset into two halves for training and testing, and report average performance over
10 random splits. 
\vspace{0.6em}\\
\textbf{KTH TIPS 2a
dataset}\footnote{http://www.nada.kth.se/cvap/datasets/kth-tips/}~\cite{Caputo:ICCV05}
is a dataset for material categorization. It contains 11 materials \eg\ cork,
wool, linen, with images of four samples for each material. The samples were
photographed at 9 scales, 3 poses and 4 different illumination conditions. All
these variations make it an extremely challenging dataset. We use the standard
protocol~\cite{Chen:PAMI10,Caputo:ICCV05} and report the average performance
over the 4 runs, where every time all images of one sample are taken for test
while the images of the remaining 3 samples are used for training.

\vspace{0.6em}
We now analyze the performance of the proposed LHS \vs the LBP/LTP based image representations.
Tab.\ \ref{tab:exp} (col.\ 1 and 2) gives the results for the different methods on the
texture datasets. On the texture recognition experiment, \ie when a sample seen on training is
presented for testing with scale and rotation changes, we achieve a near perfect accuracy of 99.5\%.
Our best method outperforms the best LBP and LTP baselines by 12.2\% and 4.5\% respectively. We thus
conclude that data-adaptive encoding, using higher-order statistics, of local neighborhoods is
advantageous when compared to fixed quantization and heuristics as used in LBP and LTP
representations. The high accuracy achieved on the texture recognition dataset, Brodatz-32, leads us
to conclude that texture recognition, under the presence of rotation and
scale variations, can be done almost perfectly. 

On the more challenging KTH TIPS 2a dataset for texture categorization, the best performance we
obtain is far from saturated at 73\%. LHS performs better than LBP and LTP baselines by 3.2\% and
1.7\% respectively. KTH TIPS 2a dataset has much stronger variations in scale, illumination
conditions, pose, \etc than the Brodatz dataset and the experiment is of texture categorization of
unseen sample \ie the test images are of a sample not seen in training. LHS again outperforms
LBP/LTP on the task of texture categorization. More recently better results have been reported on
the task of texture categorization. It has been demonstrated that standard object image
classification pipeline of Fisher Vectors \cite{Jaakkola:NIPS99,SanchezIJCV2013} with dense SIFT
\cite{Lowe:IJCV04} when applied to texture categorization \cite{CimpoiCVPR2014} achieves excellent
results. We note that such features are of much higher complexity than the proposed LHS. We analyze
LHS \wrt such features in Sec.~\ref{sec:lfw_sup}, albeit on the task of face verification. Also, it
has been shown that representations learnt for image classification tasks, using large amounts of
external data, transfer successfully to texture recognition as well \cite{CimpoiCVPR2014}. While
such methods are quite interesting, they are not directly comparable to the proposed method.

\begin{table*}[t]
\begin{center}
\vspace{-1em}
\subfigure[Rectangular sampling (8-pixel neighborhood)]{
\begin{tabular}{lcccc}
\hline
       &  \; \; Brodatz--32 \; \; & KTH TIPS 2a & \; JAFFE E1 \; & JAFFE E2  \\
\hline\hline                           
LBP baseline &        87.2 $\pm$ 1.5  &        69.8 $\pm$ 6.9  &        86.9 $\pm$ 2.6  &        56.5 $\pm$ 21.0  \\ 
LTP baseline &        95.0 $\pm$ 0.8  &        69.3 $\pm$ 5.3  &        93.6 $\pm$ 1.8  &        57.2 $\pm$ 16.3  \\ 
LHS (proposed)   &\textbf{99.3 $\pm$ 0.3} &\textbf{71.7 $\pm$ 5.7} &\textbf{95.6 $\pm$ 1.7} &\textbf{64.6 $\pm$ 19.2} \\
\hline
\end{tabular}
}
\subfigure[Circular sampling (bilinear interpolation for diag.\ neighbproposed)]{
\begin{tabular}{lcccc}
\hline
       &  \; \; Brodatz--32 \; \; & KTH TIPS 2a & \; JAFFE E1 \; & JAFFE E2  \\
\hline\hline                           
LBP baseline &        87.3 $\pm$ 1.5  &        69.8 $\pm$ 6.7  &        94.3 $\pm$ 2.1  &        61.8 $\pm$ 24.1  \\ 
LTP baseline &        94.9 $\pm$ 0.8  &        71.3 $\pm$ 6.3  &        95.1 $\pm$ 1.8  &        60.6 $\pm$ 20.8  \\ 
LHS (proposed)   &\textbf{99.5 $\pm$ 0.2} &\textbf{73.0 $\pm$ 4.7} &\textbf{96.3 $\pm$ 1.5} &\textbf{63.2 $\pm$ 16.5} \\
\hline
\end{tabular}
}
\end{center}
\vspace{-1.5em}
\caption{
Results (avg.\ accuracy and std.\ dev.) on the different datasets.
}
\label{tab:exp}
\end{table*}

\subsection{Facial analysis}
\label{sec:exp_jaffe}
\noindent \textbf{Japanese Female Facial Expressions
(JAFFE)}\footnote{http://www.kasrl.org/jaffe.html} \cite{Lyons:AFGR98} is a
dataset for facial expression recognition. It contains 10 different females
expressing 7 different emotions \eg sad, happy, angry.  We perform expression
recognition for both known persons, like earlier works \cite{Liao:ICIP06},
and for unknown person. In the first (experiment E1), one image per expression for each
person is used for testing while remaining ones and used for training.  Thus, the person being
tested is present (different images) in training. In the second (experiment E2), all images of one
person are held out for testing while the rest are used for training. Hence, there are no images of
the person being tested in the training images, making the task more challenging. For both cases, we
report the mean and standard deviation of average accuracies of 10 runs.
\begin{figure}[t]
\begin{center}
\includegraphics[width=\columnwidth, trim=0 370 300 0, clip]{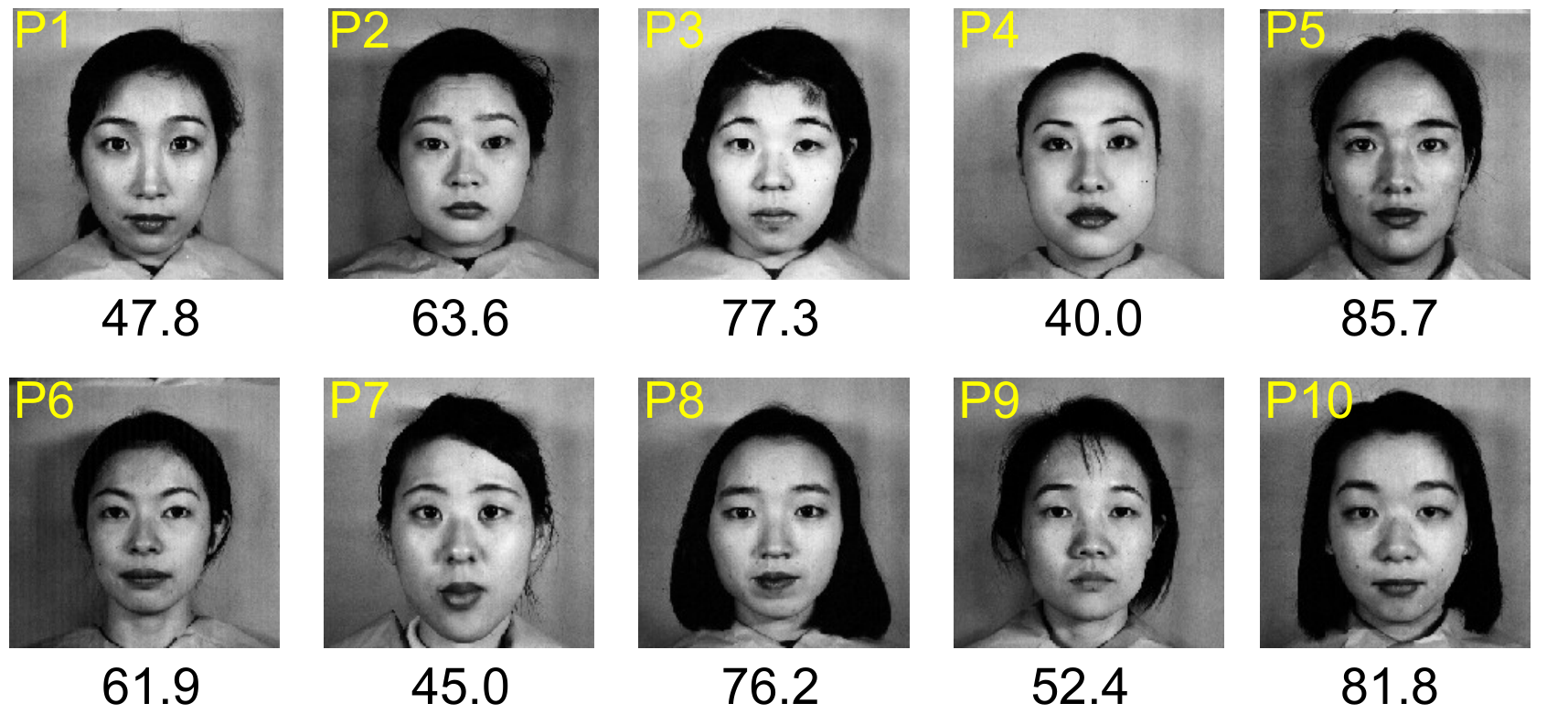}
\vspace{-1em} 
\caption{The images of the 10 persons in the neutral expression. The number
below is the categorization accuracy for all 7 expressions for the person (see
Sec.\ \ref{sec:exp_jaffe}).}
\label{fig:jaffe_e2}
\vspace{-1em} 
\end{center}
\end{figure}

\begin{figure}[t]
\centering
\includegraphics[width=\columnwidth, trim=0 240 0 0, clip]{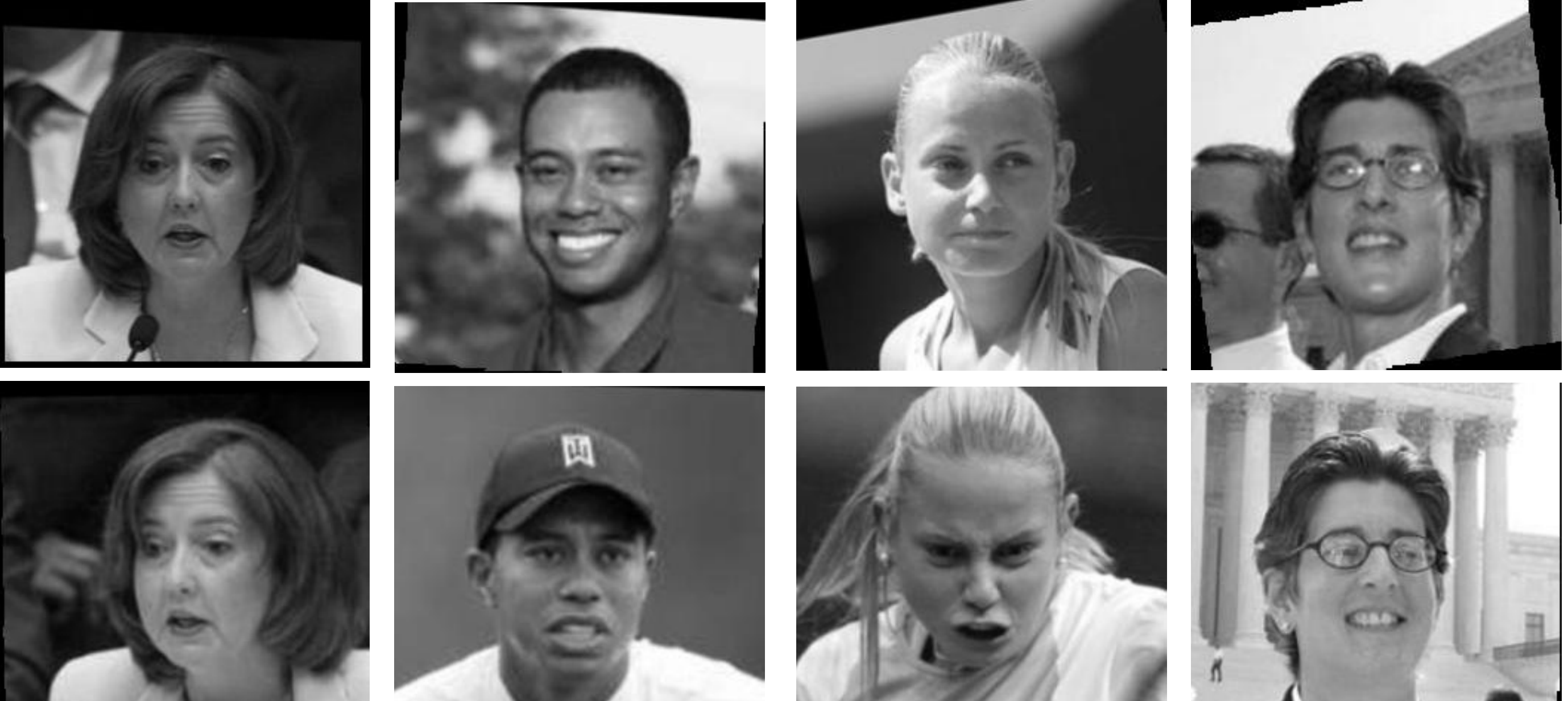} 
\caption{
Example image pairs from the LFW \cite{LFWTech,Wolf:ACCV09} dataset. Note the large
variation in appearance due to different pose, expression, illumination, accessories etc.
}
\label{fig:lfw_egs}
\end{figure}

\vspace{0.6em}
Tab.\ \ref{tab:exp} (col.\ 3 and 4) gives the performance of the different methods on the expression
categorization task. On the first experiment (E1) we obtain very high accuracies as the task is of
recognition of expressions, from a never seen image, of a person who was already seen at training. The
proposed LHS again outperforms LBP and LTP based representation by 2\% and 1.2\%, respectively. On
the more challenging second experiment (E2), \ie when the test subject was not seen during training,
we see that the accuracies are much less than E1. LHS again outperforms the best LBP and LTP
accuracies by 2.8\% and 4\% respectively. Fig.\ \ref{fig:jaffe_e2} shows one image of each of the 10
persons in the dataset along with the expression recognition accuracy for that person. This dataset
has highly variable intra-person differences \ie for some individuals different expression images
are close while for others they are very different. This results in very different accuracies for
the different persons and hence high standard deviation, for all the methods. We conclude that LHS,
owing to more accurate description of local pixel neighborhoods, is able to perform better than the
LBP/LTP based image description for the task of facial expression categorization on the JAFFE
dataset.
\vspace{0.6em}\\
\textbf{Labeled Faces in Wild (LFW)}~\cite{LFWTech} is a popular dataset for face verification by
unconstrained pair matching. Face verification is the task where two face
images are given and the system has to predict whether they are of the same
person or not, with the possibility that the(those) person(s) might not have been seen at training.
Hence, it is different from face recognition, where the system has to recognize a person already seen
at training. It stresses the system to find characteristics which are general and make the faces
similar or not, rather than characteristics which are specific to a known set of persons. LFW contains
13,233 face images of 5749 different individuals of different ethnicity, gender, age, \etc It is an
extremely challenging dataset and contains face images with large variations in pose, lighting,
clothing, hairstyles, \etc (Fig.~\ref{fig:lfw_egs} shows example pairs from the dataset). LFW
dataset is organized into two parts: `View 1' is used for training and validation (\eg for choosing
the parameters) while `View 2' is only for final testing and benchmarking. In our setup, we follow
the specified training and evaluation protocol. We use the, publicly available, aligned version of
the faces as provided by Wolf \etal
\cite{Wolf:ACCV09}\footnote{http://www.openu.ac.il/home/hassner/data/lfwa/}.

\begin{figure*}[t]
\begin{center}
\includegraphics[width=0.39\textwidth, trim=0 350 440 0, clip]{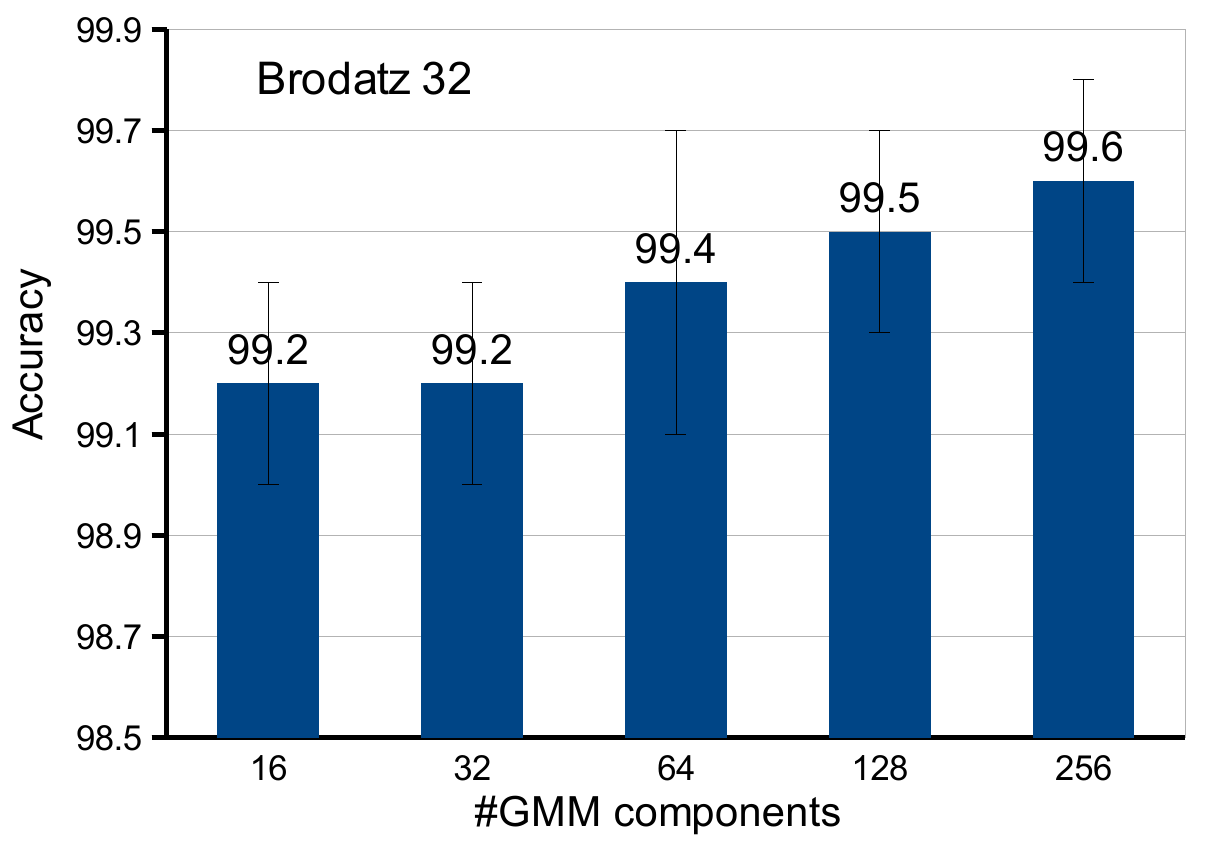} \hspace{3em}
\includegraphics[width=0.40\textwidth, trim=350 350 80 0, clip]{tex_gmm_comp}
\vspace{-0.5em}
\caption{The accuracies of the method for different number of GMM components for Brodatz (left) and KTH TIPS 2a (right)
dataset (see Sec.\ \ref{sec:params})}
\vspace{-1em}
\label{fig:tex_gmm_comp}
\end{center}
\end{figure*}

\begin{table*}[t]
\renewcommand{\arraystretch}{1.1}
\newcolumntype{L}[1]{>{\raggedright\let\newline\\\arraybackslash\hspace{0pt}}m{#1}}
\newcolumntype{C}{>{\centering\arraybackslash}p{25em}}
\begin{center}
\begin{tabular}{CC}
\subfigure[Brodatz--32]{
\begin{tabular}{|l|c|L{11em}|}
\hline
Method &Acc. & Remark \\ 
\hline \hline
Jalba \etal \cite{JalbaPR2004} & 93.5 & Morphological hat-transform \\
\hline
Urbach \etal \cite{Urbach:PAMI07} & 96.5 & Connected shape size pattern spectra \\ 
\hline
Ojala \etal \cite{OjalaPR2001} &96.8 & Distributions of signed gray level differences\\
\hline
Chen \etal \cite{Chen:PAMI10} &97.5 & Weber law feat.\ + $k$-NN  \\ 
\hline
LHS (proposed)        &\textbf{99.3} & \\
\hline
\end{tabular}
}
&
\subfigure[JAFFE]{
\begin{tabular}{|l|c|L{11em}|}
\hline
Method &Acc. & Remark \\ 
\hline \hline
Shan \etal \cite{Shan:IVC09} &81.0 & LBP based\\ 
\hline
Guo \etal \cite{GuoCVPR2003} &91.0 & Gabor filters + feat.\ selection \\ 
\hline
Lyons \etal \cite{LyonsPAMI1999} &92.0 & Gabor filters + Linear Discriminant Analysis \\ 
\hline
Feng \etal \cite{Feng:PRAI05}&93.8 & LBP + Linear programming\\ 
\hline
LHS (proposed)              &\textbf{95.6} &  \\
\hline
\end{tabular}
}
\\
\subfigure[KTH TIPS 2a]{
\begin{tabular}{|l|c|L{11em}|}
\hline
Method &Acc. & Remark \\ 
\hline \hline
Chen \etal \cite{Chen:PAMI10}    &64.7 & Weber law feat.\ + $k$-NN\\ 
\hline
Caputo \etal \cite{Caputo:ICCV05} &71.0 & 3 sc.\ LBP, nonlin.\ SVM\\ 
\hline
LHS (proposed)                   &73.0 & \\
\hline
DeCAF \cite{CimpoiCVPR2014}&78.4 & Large amount of labeled external data\\
\hline
SIFT-FV \cite{CimpoiCVPR2014} &\textbf{82.2} & Higher complexity, see \S~\ref{sec:lhs_vs_high}\\
\hline
\end{tabular}
}
&
\subfigure[LFW (aligned, unsupervised)]{
\begin{tabular}{|l|c|L{9em}|}
\hline
Method &Acc. & Remark \\ 
\hline \hline
{\small Javier \etal}\cite{Javier:EURASIP09} &69.5 {\footnotesize$\pm$0.5}& LBP with $\chi^2$ dist. \\ 
\hline
Seo \etal \cite{Seo:IFS11}&72.2 {\footnotesize$\pm$0.5}& Locally Adaptive Regression Kernel \\ 
\hline
LHS {\small (proposed)} &73.4 {\footnotesize$\pm$0.4}& \\
\hline
LQP \cite{HussainBMVC2012} &75.3 {\footnotesize$\pm$0.8}& Higher complexity\\
\hline
PAF \cite{YiCVPR2013} &\textbf{87.8 {\footnotesize$\pm$0.5}}& External data for pose correction\\
\hline
\end{tabular}
}
\end{tabular}
\caption{
Comparison with current methods with comparable experimental setup (reports accuracy, see Sec.\ \ref{sec:stateofart}).
}
\label{tab:stateofart}
\end{center}
\end{table*}

We first report results in the restricted unsupervised task of the LFW dataset, \ie (i) we use
strictly the data provided without any other data from any other source and (ii) we do not utilize
class labels while obtaining the image representation. This task evaluates the information contained
in the features without help from any supervised modifications. We will provide results later in the
supervised setting in Sec.~\ref{sec:lfw_sup} where we will demonstrate that, combined with supervised
learning, LHS give a very attractive trade-off between performance and speed \wrt the
state-of-the-art methods.

We center crop the face images to 150$\times$80 and resize them to 70$\times$40 pixels. We then
compute the features with a 7$\times$4 grid, of 10$\times$10 pixels cells, overlayed on the image.
We compute the LHS representations for each cell separately and compute the similarity between image
pairs as the mean of L2 distances between the representations of corresponding cells. We classify
image pairs into same or not same by thresholding on their similarity. We choose the testing
threshold as the one which gives the best classification accuracy on the training data. 

LHS gives an accuracy of 73.4\% with a standard error on the mean of 0.4\%. This is a competitive
performance in the unsupervised setting for the dataset. Unlike other methods, it neither uses
external data \eg as by PAF \cite{YiCVPR2013}, nor does it do feature-specific post-processings \eg
as by LQP \cite{HussainBMVC2012}. We compare with other existing approaches, including those based
on LBP in Sec.\ \ref{sec:stateofart}. Also, we show in Sec.\ \ref{sec:lfw_sup} that LHS is one of the
best performing methods, among methods of similar low complexity, in the supervised face verification setting.

\subsection{Effect of sampling and number of components}
\label{sec:params}
Table \ref{tab:exp} shows the results with (a) rectangular 3$\times$3 pixel neighborhood and (b)
LBP/LTP like circular sampling of 8 neighbors with the diagonal neighbor values bilinearly
interpolated. Performance on the Brodatz-32 dataset is similar for both the samplings while that for
KTH and JAFFE datasets differ. In general, the circular sampling performs better for all the
datasets. We note that the variations in, and hence difficulty of, the Brodatz-32 dataset is much less
compared to the other two datasets and hence images in Brodatz-32 dataset are possibly well
represented by either of the two samplings. Thus, we conclude that, in general, circular sampling is
to be preferred as it performs better on most of the datasets and generates more discriminative
statistics.

Fig.~\ref{fig:tex_gmm_comp} shows the performance on the two texture datasets for different number
of mixture model components. The length of the vector, and hence the space and time complexity of
the method, varies proportional to the number of components in the GMM. Relatively higher number of
components leads to a higher likelihood, \ie a better fit to the data, and hence a better
description of the space but also leads to vectors which are longer to compute and store. On this
trade-off of size and accuracy of description, we observe that the performance, for both texture
datasets, improves with the number of components and saturates at 128. On the Brodatz-32 dataset,
LHS is able to give more than 99\% accuracy with just 16 components, highlighting the relatively
easier nature of this dataset. While for the KTH dataset, performance improves significantly when
the number of components increase from 16 to 128 (by absolute 6.8\%). KTH is significantly more challenging than the
Brodatz-32 dataset and hence requires more accurate and costly descriptors computed from larger
number of mixture components.

\begin{table*}[t]
\begin{minipage}[c]{0.48\textwidth}
\centering
\renewcommand{\arraystretch}{1.035}
\begin{tabular}{c|c|c|c}
\multicolumn{4}{c}{Supervised, unrestricted, label free outside data} \\
\hline
\#Gauss.\ & \multicolumn{2}{c|}{Dimension}& ROC-EER \\ 
   $(k)$    & org.\ ($d_0$) & proj.\ ($d$) & Accuracy \\ 
\hline
\hline
\multirow{3}{*}{4} & \multirow{3}{*}{1792} &   32  &        85.73 $\pm$ 0.17\\
                   &                       &   64  &        86.37 $\pm$ 0.19\\
                   &                       &  128  &\textbf{86.60 $\pm$ 0.17}\\
\hline
\multirow{3}{*}{8} & \multirow{3}{*}{3584} &   32  &        86.47 $\pm$ 0.17\\
                   &                       &   64  &        87.37 $\pm$ 0.14\\
                   &                       &  128  &\textbf{87.60 $\pm$ 0.14}\\
\hline
\multirow{3}{*}{16}& \multirow{3}{*}{7168} &   32  &        87.43 $\pm$ 0.17\\
                   &                       &   64  &        87.57 $\pm$ 0.17\\
                   &                       &  128  &\textbf{88.13 $\pm$ 0.15}\\ 
\hline
\multirow{3}{*}{24}& \multirow{3}{*}{10752}&   32  &        87.63 $\pm$ 0.17\\ 
                   &                       &   64  &        87.93 $\pm$ 0.20\\
                   &                       &  128  &\textbf{88.27 $\pm$ 0.17}\\
\hline
\multirow{3}{*}{32}& \multirow{3}{*}{14336}&   32  &        87.47 $\pm$ 0.20\\ 
                   &                       &   64  &\textbf{88.03 $\pm$ 0.16}\\
                   &                       &  128  &\textbf{87.97 $\pm$ 0.14}\\
\hline 
\end{tabular}
\caption{
Results of proposed LHS on the Labeled Faces in the Wild (LFW) \cite{LFWTech} dataset
for different parameter settings.
}
\label{tab:lfw_exp}
\end{minipage}
\hfill
\begin{minipage}[c]{0.48\textwidth}
\centering
\begin{tabular}{c|c}
\multicolumn{2}{c}{Methods with similar complexity} \\
\hline
Method & Accuracy \\
\hline
\hline
LBP + ITML \cite{Taigman:BMVC09} & 85.1 $\pm$ 0.6 \\
LBP + PLDA \cite{Li:PAMI2012} & 87.3 $\pm$ 0.6 \\
LHS + JML (proposed) \; & \textbf{88.2 $\pm$ 0.2} \\
\hline 
\multicolumn{2}{c}{ } \\[-5pt]
\multicolumn{2}{c}{Methods with multiple feats/higher complexity} \\
\hline
Method & Accuracy \\
\hline 
\hline 
comb.\ LDML-MkNN \cite{Guillaumin:ICCV2009} & 87.5 $\pm$ 0.4 \\
comb.\ PLDA \cite{Li:PAMI2012} & 90.1 $\pm$ 0.5 \\
SIFT-FV \cite{SimonyanBMVC2013} & 93.0 $\pm$ 1.1 \\
High dim LBP \cite{ChenCVPR2013} & 93.2 $\pm$ 1.1 \\
LBP + LHS (proposed)  & 89.0 $\pm$ 0.1 \\
SIFT-FV + LHS (proposed)  & \textbf{93.5 $\pm$ 0.2} \\
\hline
\multicolumn{2}{c}{ } \\[-5pt]
\multicolumn{2}{c}{Methods using large amts of external labeled data} \\
\hline
Method & Accuracy \\
\hline
\hline
High dim LBP \cite{ChenCVPR2013} & 95.2 $\pm$ 1.1 \\
Deep learning \cite{SunCVPR2014,TaigmanCVPR2014} & \textbf{97.4 $\pm$ 0.3} \\
\hline
\end{tabular}
\caption{
Comparison with existing works on the Labeled Faces in the Wild (LFW) \cite{LFWTech}
dataset--unrestricted and supervised setting.  
}
\label{tab:lfw_soa}
\end{minipage}
\end{table*}

\subsection{Comparison with existing methods}
\label{sec:stateofart}
Table \ref{tab:stateofart} shows the performance of the proposed LHS along with existing methods. On the
Brodatz dataset we outperform all methods and to the best of our knowledge report, near perfect,
state-of-the-art performance. On the JAFFE dataset, as well, we achieve the best results reported
till date. 
On the KTH dataset, Chen \etal \cite{Chen:PAMI10} report an accuracy of 64.7\% using their Weber law
descriptors (WLD) with KNN classifier. Caputo \etal\ \cite{Caputo:ICCV05} report 71.0\% for their
3-scale LBP and \emph{non-linear} chi-squared radial basis function kernel based SVM classifier. In
comparison we use linear classifiers which are not only fast to train but also need only a vector
dot product at test time (\cf kernel computation with support vectors which is of the order of
number of training features). Note Caputo \etal obtain their best results with multi scale features
and a complex decision tree (with non-linear classifiers at every node). We expect our features to
outperform their features with similar complex classification architecture. The higher complexity
Fisher vectors with SIFT features (SIFT-FV) \cite{CimpoiCVPR2014} achieves substantially better on
the KTH dataset (82.2\% \vs 73.0\%), however they are orders of magnitude longer and slower than the
proposed method. We provide space and time comparisons with the proposed method and SIFT-FV method
below Sec.~\ref{sec:lhs_vs_high}, Tab.~\ref{tab:lhs_vs_high}, with images from LFW dataset without
loss of generality.

Tab.\ \ref{tab:stateofart}(d) reports accuracy rates of our method and those of competing
unsupervised methods\footnote{For more results, see webpage
\url{http://vis-www.cs.umass.edu/lfw/results.html}} on LFW dataset. Our method outperforms
the LBP  baseline (LBP with $\chi^2$ distance)~\cite{Javier:EURASIP09} by 3.9\% and gives
1.2\% better performance than Locally Adaptive Regression Kernel (LARK) features of
\cite{Seo:IFS11}. The better performance of our features, compared to the LBP baseline and fairly
complex LARK features, on this difficult dataset once again underlines the fact that
local neighborhood contains a lot of discriminative information. It also demonstrates the
representational power of our features, which are successful in encoding the information 
missed by other methods. More recent works have reported higher performances \eg Local Quantized
Patterns (LQP) \cite{HussainBMVC2012} achieves 75.3\% without any postprocessing and gain even higher
when postprocessed with whitened PCA and compared with cosine similarity. However, LQP have higher
complexity than LHS and hence gain 2\%. While the current LHS features are only 3584 dimensional,
the LQP features are 36000 dimensional \ie 10$\times$ longer. Pose Adaptive Filters (PAF)
\cite{YiCVPR2013} use external data to learn pose robust features using 3D fitting of faces and
achieve substantially
more. This underlines the fact that the dataset has very challenging pose variations, correcting
which will arguably improve the performance of the proposed LHS features as well. Since they use
external data while the proposed method does not, their performance is not directly comparable to
that of the proposed method. Also, adding pose robustness with additional effort, \eg 3D fitting of
face and using external data, is another challenging problem in itself and has not been explored
further in this work.

Thus, we conclude, the proposed method is capable of achieving competitive results while being
computationally simple and efficient.

\subsection{LHS with supervised discriminative metric learning}
\label{sec:lfw_sup}
We now provide results of the proposed Local Higher-order Statistics (LHS) features with supervised
discriminative metric learning (ML) on the challenging Labeled Faces in the Wild (LFW)
\cite{LFWTech} dataset. We show that when used with such supervised ML, which can be
equivalently seen as a projection to a lower dimensional discriminative subspace (see
Sec.~\ref{sec:ml}), the LHS features can obtain very high performance while being much more
efficient than the competition.

We operate in the \emph{`Supervised, unrestricted, label-free outside data'} protocol.
Tab.~\ref{tab:lfw_exp} gives the performance of LHS for different values of the parameters. We see
that the increasing the number of Gaussian components steadily increases the performance from $k=4$
to $k=24$ by a little less than 2\% absolute while beyond that the results seem to saturate.
Similarly, for a fixed number of Gaussian components, increasing the projection dimension increases the
results but with a pronounced diminishing returns effect.

It is quite interesting to note these performances in the context of existing methods.
Tab.~\ref{tab:lfw_soa} shows the performance of LHS \wrt state-of-the-art methods on LFW dataset.
LFW achieves the best results among the features in the low complexity regime, and competitive
results among features with high complexity or methods that combine multiple features. In particular
our own implementation of Local Binary Patterns (LBP) using the (default parameters of the) 
\texttt{vlfeat} library \cite{vlfeat} gives 86.2\% with a feature dimension of 7k. Compared to this
LHS with only 1k dimensions gives 86.6\% (Tab.~\ref{tab:lfw_exp}) and that with 10k dimension gives
88.3\%. When combined with LBP the performance increases to 89.0\%. Our implementation of Fisher
vectors with dense SIFT features gives 92.9\% (compared to 93.0\% reported in
\cite{SimonyanBMVC2013}), and when combined with LHS the performance improves to 93.5\%, which is a
modest improvement over the state-of-the-art in the \emph{`Supervised, unrestricted, label-free
outside data'} protocol\footnote{For more results, see webpage
\url{http://vis-www.cs.umass.edu/lfw/results.html}}.
Thus, we conclude that LHS features are competitive in the low complexity domains and are
complementary to the high complexity features for supervised face verification on LFW.
In the next section we discuss their time and space benefit over the high complexity features.

\begin{table}[t]
\centering
\begin{tabular}{c|c|c|c|c}
\hline
& \multicolumn{2}{c|}{Space} & \multicolumn{2}{c}{Time} \\
Method & dim.\ & reduction & \; \; ms \; \; & speedup \\
\hline \hline
SIFT-FV              & 67584 &Ref. & 2400* & Ref.\\
\multirow{5}{*}{LHS} & 1792  & 38$\times$ & 13 & 185$\times$\\ 
                     & 3584  & 19$\times$ & 15 & 160$\times$\\
                     & 7168  &  9$\times$ & 19 & 126$\times$ \\
                     & 10752 &  6$\times$ & 22 & 109$\times$ \\
                     & 14336 &  5$\times$ & 25 &  96$\times$ \\
\hline                                       
\end{tabular}
\caption{
The space and time complexity comparison between proposed LHS the FV method. (*) The time for the
best performing configuration in \cite{SimonyanBMVC2013}, \ie step size 1, is interpolated from the
time reported for step size 2 (0.6s). Our implementation of fisher vectors takes similar time, see
Sec.~\ref{sec:lhs_vs_high}
}
\label{tab:lhs_vs_high}
\end{table}

\subsection{Time and space complexity of LHS}
\label{sec:lhs_vs_high}
The proposed LHS features are very compact and efficient to compute.  Compared to one of the
state-of-the-art systems for face verification \cite{SimonyanBMVC2013} they are about two orders of
magnitude faster and an order of magnitude smaller.  Tab.~\ref{tab:lhs_vs_high} gives the space and
computation time comparison of the LHS features \wrt Fisher vectors with SIFT features (SIFT-FV)
\cite{SimonyanBMVC2013}. The experiments were run on a server with 2.67 GHz Intel Xeon processor 
running Ubuntu 14.04 and all the data was loaded in RAM for timing the computations. The times
reported are for a single threaded program using one core.

The best performing LHS features are 10,752 dimensional and take 22 ms to
compute compared to 67,584 for SIFT-FV which amounts to a space saving of 6$\times$ and speedup of
109$\times$; while the most lightweight LHS configuration tested is 38$\times$ smaller and
185$\times$ faster than SIFT-FV. Ignoring the offline training time, which is $O(d_0^2)$, and
considering only the online testing times, the best performance is reached when the image pairs are
horizontally flipped and the distance between the four combinations are averaged. Hence, for
comparing a face pair, features for 4 images need to be calculated \ie Fisher vectors take 9.6s
while the proposed LHS take only 88ms, both on a single core of a modern CPU. Such advantages come
with a drop in performance, but might be essential for time and space critical applications \eg in
embedded systems. They might also be used in a cascade system where the efficient LHS features are
used to tackle the easy decisions while delegating the tougher examples to the higher complexity
features, thereby reducing the average time over several comparisons.

We note that, our implementation of LHS is in unoptimized C/C++, called via the MEX interface of
MATLAB. Arguably it can be improved substantially, in particular, by tuning/approximating the GMM
posterior probability estimation, which involves costly exponential operations.

\section{Conclusions}
We have presented a model that captures higher-order statistics of small local neighbohoods to
produce a highly discriminative representation of the images. Our experiments, on two challenging
texture datasets and two challenging facial analysis datasets, validate our approach and show that
the proposed model encodes more local information than the competing methods and achieves
competitive results. Two of the datasets we used, one each for textures and faces, were relatively
simpler while two other were more difficult. The results on the simpler datasets served to
demonstrate that the method is capable of having a richer appearance descriptor compared to existing
methods.  While on the more challenging case, we showed with experiments on the supervised task of
face verification on the challenging Labeled Faces in the Wild (LFW) dataset that the proposed
method achieves best results for low complexity features and is complementary to the high dimensional
features. When combine with the state-of-the-art method it improves the performance to establish a
new state-of-the-art on the LFW dataset when no external labeled data is used. Compared to the best
method the proposed method is two orders of magnitude faster to compute and an order of magnitude
compact making it a very appropriate choice for low complexity devices \eg embedded systems.

While the current state-of-the-art systems, based on deep networks trained with large amounts of
external data \cite{DuongCVPR2015, HuangCVPR2012, HuCVPR2014, SchroffCVPR2015, SunCVPR2014,
SunCVPR2015, TaigmanCVPR2014, TaigmanCVPR2015, KanCVPR2014, ZhuICCV2013} have gained, the proposed
method is still relevant due to its speed -- we could use it as an initial low complexity stage of a
cascade based system. Also, in very low complexity/cost systems the size of the model might also
limit the use of deep networks and make the proposed method relevant.

\vspace{1em}
\noindent \textbf{Acknowledgments.}
The authors acknowledge support from the PHYSIONOMIE project, grant number ANR-12-SECU-0005-01. 

\section*{References}
\bibliography{biblio}
\vspace{2em}

\parpic{\includegraphics[width=1in,height=1.25in,clip,keepaspectratio]{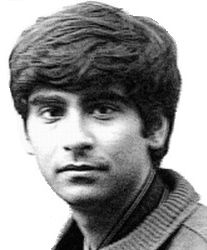}}
{\small
\noindent {\bf Gaurav Sharma}
is currently at the Max Planck Institute for Informatics, Germany. He holds an Integrated Master of
Technology (5 years programme) in Mathematics and Computing from the Indian Institute of Technology
Delhi (IIT Delhi) and a PhD in Applied Computer Science from INRIA (LEAR team) and the Universit\'e
de Caen Basse-Normandie, France. His primary research interest lies in Machine Learning applied to
Computer Vision tasks such as image classification, object recognition and facial analysis.
}

\vspace{2em}
\parpic{\includegraphics[width=1in,height=1.25in,clip,keepaspectratio]{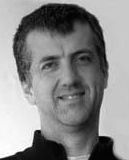}}
{\small
\noindent {\bf Fr\'ed\'eric~Jurie}
is a professor at the French Universit\'e de Caen Basse-Normandie (GREYC - CNRS UMR6072) and an
associate member of the INRIA-LEAR team. His research interests lie predominately in the area of
Computer Vision, particularly with respect to object recognition, image classification and object
detection.
}
\end{document}